\documentclass[10pt]{article} 
\usepackage[preprint]{tmlr}

\usepackage{amsmath,amsfonts,bm}

\def\eqref#1{equation~\ref{#1}}

\def\1{\bm{1}}

\DeclareMathAlphabet{\mathsfit}{\encodingdefault}{\sfdefault}{m}{sl}
\SetMathAlphabet{\mathsfit}{bold}{\encodingdefault}{\sfdefault}{bx}{n}

\usepackage[utf8]{inputenc} 
\usepackage[T1]{fontenc}    
\usepackage{hyperref}       
\usepackage{url}            
\usepackage{booktabs}       
\usepackage{amsfonts}       
\usepackage{nicefrac}       
\usepackage{microtype}      
\usepackage{xcolor}         

\usepackage{amsmath}
\usepackage{amssymb}
\usepackage{mathtools}
\usepackage{amsthm}
\usepackage{algorithmic}
\usepackage{algorithm}
\usepackage{float}
\usepackage{tikz-cd}
\usepackage{natbib}
\usepackage{relsize}
\usepackage{scalerel}
\theoremstyle{plain}
\newtheorem{theorem}{Theorem}[section]

\newtheorem{lemma}[theorem]{Lemma}

\theoremstyle{definition}
\newtheorem{definition}[theorem]{Definition}

\theoremstyle{remark}

\DeclareMathOperator{\Col}{Col}
\DeclareMathOperator{\CSoftmax}{Cell-Softmax}
\DeclareMathOperator*{\concat}{\scalerel*{\Vert}{\sum}}

\usepackage{amssymb,amsmath,stackengine}
\stackMath
\usepackage{ifthen}

\counterwithin*{equation}{section}
\usepackage{fancyhdr}

\title{CW-CNN \& CW-AN: Convolutional Networks and Attention Networks for CW-Complexes}

\author{\name Rahul Khorana \email rahul.khorana@berkeley.edu \\
      \addr Department of Computer Science\\
      University of California, Berkeley
    }

\def\month{MM}  
\def\year{YYYY} 
\def\openreview{\url{https://openreview.net/forum?id=XXXX}} 

\begin{document}

\maketitle

\begin{abstract}
We present a novel framework for learning on CW-complex structured data points. Recent advances have discussed CW-complexes as ideal learning representations for problems in cheminformatics. However, there is a lack of available machine learning methods suitable for learning on CW-complexes. In this paper we develop notions of convolution and attention that are well defined for CW-complexes. These notions enable us to create the first Hodge informed neural network that can receive a CW-complex as input. We illustrate and interpret this framework in the context of supervised prediction.
\end{abstract}

\section{Introduction}

\subsection{Complexes}
Succinctly, a cell complex is an object in a category obtained by successively gluing together cells using pushouts. More formally, \citet{whitehead1949combinatorial} defined them in the following way. 
\begin{definition}
\label{def:CellComplex}
A cell complex $K$, or alternatively a \textit{complex}, is a Hausdorff space which is the union of disjoint open cells $e,e^{n},e_{i}^{n}$ subject to the condition that the closure $\bar{e}^{n}$ of each $n$-cell, $e^{n} \in K$ is the image of a fixed $n$-simplex in a map $f:\sigma^{n} \to \bar{e}^{n}$ such that 
\begin{itemize}
    \item [(1)] $f | \sigma^{n} - \partial \sigma^{n}$ is a homeomorphism onto $e^{n}$
    \item [(2)] $\partial e^{n} \subset K^{n-1}$, where $\partial e^{n} = f\partial \sigma^{n} = \bar{e}^{n} - e^{n}$ and $K^{n-1}$ is the $(n-1)$-section of $K$ consisting of all the cells whose dimensionalities do not exceed $n-1$.
\end{itemize}
\end{definition}
A CW-complex is a cell complex that has the weak topology and is closure finite. A complex $K$ is said to be closure finite if and only if $K(e)$ is a finite subcomplex, for every cell $e \in K$. We say $K$ has the weak-topology if and only if a subset $X \subset K$ is closed provided $X \cap \bar{e}$ is closed for each cell $e \in K$. 
To construct a CW-complex, we inductively glue cells together. More formally, \citet{hatcher2002algebraic} describes how we construct a finite CW-complex $X$ as follows.
Initially, we start with a collection of zero cells $X^{0} = \{e_{i}^{0}\}_{i=0}^{N}$. $X^{0}$ is called the $0$-skeleton.
Then, for all $j\in\{1,\ldots,n\}$ we take a collection of $j$-cells $\{e_{i}^{j}\}_{i=0}^{N}$ and glue their boundaries to points in the $j-1$ skeleton using continuous attaching maps $\phi_{i}^{j}: \partial e_{i}^{j} \to X^{j-1}$. Each $j$-cell is a topological space. Essentially, a CW-complex is constructed by taking a union of a sequence of topological spaces $\o = X^{-1} \subset X^{0} \subset X^{1} \subset \cdots$ such that each $X^{j}$ is obtained by attaching $j$-cells to $X^{j-1}$. In the language of category theory, we often think of the topology on finite CW-complex $X$ as the direct limit of the diagram $X^{-1} \hookrightarrow X^{0} \hookrightarrow X^{1} \hookrightarrow \cdots \hookrightarrow X^{k}$ for some $k \in \mathbb{N}$. CW-complexes generalize the notion of a graph. A $1$-dimensional CW-complex is a regular graph without loops. Moreover, every topological space is weakly homotopy equivalent to a CW-complex.
\subsection{Learning on CW-complexes}
Consider the following learning problem. Suppose we are presented with a dataset $\mathcal{D} = \{(x_i, y_i)\}_{i=1}^{n}$ where $x_i$ is a CW-complex, $y_{i} \in \mathbb{R}^{d}$, and $n, d \in \mathbb{N}$. Then, the task of learning a function $\mathcal{F}$ such that $y_i = \mathcal{F}(x_i) + \epsilon$, where $\epsilon$ is some error, naturally arises. We tackle this problem by developing a convolutional layer and attention mechanism for a CW-complex $x_i$. Essentially, we extend the work of \citet{kipfwelling2017semisupervised} to define a notion of convolution for CW-complexes. Additionally, we extend the work of \citet{GraphAttentionVelickovicBengio} to develop a notion of attention for CW-complexes.

\section{Related Work}

\subsection{Graph Neural Networks}
\citet{kipfwelling2017semisupervised} develop a semi-supervised learning framework for graphs. The authors consider a graph-based semi-supervised learning approach, where label information is smoothed over the graph by applying a form of explicit graph-based regularization. 
\begin{definition}
\label{def:convolution-graph-loss}
Let $\mathcal{G} = (V,E)$ be a graph. Let $f$ be a neural network. Let $\lambda$ be a weighting factor. Let $X$ be a matrix of node feature vectors. Let $A$ be the adjacency matrix for a graph $\mathcal{G}$. Let $D$ be the degree matrix for $\mathcal{G}$. Finally, let $\Delta = D - A$ be the unnormalized graph Laplacian. Then, $\mathcal{L} = \mathcal{L}_{0} + \lambda \mathcal{L}_{reg}$ where $\mathcal{L}_{reg} = \sum_{i,j} A_{ij} \| f(X_{i}) - f(X_{j})\|^{2} = f(X)^{\top} \Delta f(X)$. Note that $\mathcal{L}_{0}$ is the supervised loss with respect to the labeled part of the graph \citep{kipfwelling2017semisupervised}.
\end{definition}
The authors introduce a well behaved layer-wise propagation rule, and demonstrate semi-supervised classification of graphs.
\begin{definition}
\label{def:convolution-propagation-rule}
Let $\mathcal{G}$ be a graph. Let $X$ be a matrix of node feature vectors. Let $A$ be the adjacency matrix for a graph $\mathcal{G}$. Let $D$ be the degree matrix for $\mathcal{G}$. Let $\Delta = D - A$ be the unnormalized graph Laplacian. Additionally, define $\tilde{A} = A + I_{N}$ to be the adjacency matrix of $\mathcal{G}$ with self-connections, $\tilde{D}_{ii} = \sum_{j} \tilde{A}_{jj}$, $W^{(\ell)}$ be a weight matrix, and $\sigma$ a nonlinearity.
Then, we can consider a neural network $f(X,A)$ which follows the layer-wise propagation rule:
$H^{(\ell + 1)} = \sigma\left(\tilde{D}^{-\frac{1}{2}}\tilde{A}\tilde{D}^{-\frac{1}{2}}H^{(\ell)}W^{(\ell)}\right)$. Note that $H^{(\ell)} \in \mathbb{R}^{N \times M}$ is a matrix of activations in the $\ell$-th layer and $H^{(0)} = X$. Neural networks of the form $f(X,A)$ which are composed by stacking hidden layers of the form $H^{\ell}$ are called graph convolutional networks (GCN) \citep{kipfwelling2017semisupervised}.
\end{definition}

\subsection{Graph Attention Networks}
\citet{GraphAttentionVelickovicBengio} develop a notion of attention for Graphs. Let $\mathcal{G} = (\mathcal{V}, \mathcal{E})$ contains nodes $\mathcal{V} = \{1, \ldots, n\}$ and edges $\mathcal{E} \subseteq \mathcal{V} \times \mathcal{V}$, where $(j,i) \in \mathcal{E}$ denotes an edge from a node $j$ to a node $i$. We assume that every node $i \in \mathcal{V}$ has an initial representation $h_{i}^{0} \in \mathbb{R}^{d_0}$.

\begin{definition}
\label{def:GNNlayer}
A Graph Neural Network take in a set of node representations $\{h_{i} \in \mathbb{R}^{d} ~|~ i \in \mathcal{V}\}$ and the set of edges $\mathcal{E}$ as input. The layer outputs a new set of node representations $\{h_{i}^{'} \in \mathbb{R}^{d'} ~|~ i \in \mathcal{V}\}$, where the same parametric function is applied to every node given its neighbors $\mathcal{N}_{i} = \{j \in \mathcal{V} ~|~ (j,i) \in \mathcal{E}\}$: $h_{i}^{'} = f_{\theta}(h_{i}, \textsf{AGGREGATE}(\{h_{j} ~|~ j \in \mathcal{N}_{i}\}))$. The design of $f$ and $\textsf{AGGREGATE}$ is what distinguishes Graph Neural Networks \citep{BrodyAlonGAT}.
\end{definition}

\begin{definition}
\label{def:GATLayer}
A Graph Attention Network computes a learned weighted average of the representations of $\mathcal{N}_{i}$. A scoring function $e: \mathbb{R}^{d} \times \mathbb{R}^{d} \to \mathbb{R}$ computes a score for every edge $(j,i)$, which indicates the importance of the features of the neighbor $j$ to node $i$.  $e(h_i, h_j) = \text{LeakyReLU}(a^{\top} \cdot [Wh_{i} \mathbin\Vert Wh_{j}])$ where $a \in \mathbb{R}^{2d'}$, $W \in \mathbb{R}^{d' \times d}$ are learned and $\mathbin\Vert$ denotes vector concatenation. These attention scores are normalized across all neighbors $j \in \mathcal{N}_{i}$ using softmax and the attention function is defined as: $\alpha_{ij} = \text{softmax}_{j}(e(h_{i}, h_{j})) = \frac{\exp(e(h_i, h_j))}{\sum_{j' \in \mathcal{N}_{i}} \exp(e(h_i, h_{j'}))}$. Then the Graph Attention Network computes a new node representation using nonlinearity $\sigma$ as $h'_{i} = \sigma\left(\sum_{j \in \mathcal{N}_{i}} \alpha_{ij} \cdot Wh_{j}\right)$ \citep{GraphAttentionVelickovicBengio, BrodyAlonGAT}.
\end{definition}

\subsection{Gaussian Processes on Cellular Complexes}
\citet{alain2023gaussian} define the first Gaussian process on cell complexes. In doing so, the authors define the Hodge Laplacian and introduce important notation. We provide their definitions and re-use some of their notation. In order to define a path and function over a finite cellular complex or CW-complex $X$, one has to define a notion of chains and cochains.
\begin{definition}
\label{def:k-chain}
Suppose $X$ is an $n$-dimensional complex. Then, a $k$-chain $c_{k}$ for $0 \leq k \leq n$ is simply a sum over the cells:
$c_{k} = \sum_{i=1}^{N_{k}} \eta_{i} e_{i}^{k},~~\eta_{i} \in \mathbb{Z}$.
The authors show this generalizes the notion of directed paths on a graph. The set of all $k$-chains on $X$ is denoted by $C_{k}(X)$, which has the algebraic structure of a free Abelian group with basis $\{e_{i}^{k}\}_{i=1}^{N_{k}}$.
\end{definition}
\begin{definition}
\label{def:k-boundaryoperator}
Given definition \ref{def:k-chain}, the boundary operator naturally follows as $\partial_{k}: C_{k}(X) \to C_{k-1}(X)$. The operator $\partial_{k}$ maps the boundary of a $k$-chain to a $k-1$-chain. This map is linear, thereby leading to the equation:
$\partial_{k}\left( \sum_{i=1}^{N_{k}} \eta_{i} e_{i}^{k} \right) = \sum_{i=1}^{N_{k}} \eta_{i} \partial(e_{i}^{k})$.
\end{definition}
The authors then define the $k$-cochain (the dual notion of the $k$-chain) and the coboundary operator (the dual notion of the boundary operator).
\begin{definition}
\label{def:k-cochain}
Suppose $X$ is an $n$-dimensional complex. Then, a $k$-cochain on $X$ is a linear map $f: C_{k}(X) \to \mathbb{R}$ where $0 \leq k \leq n$.
$f\left(\sum_{i=1}^{N_{k}} \eta_{i} e_{i}^{k}\right) = \sum_{i=1}^{N_{k}} \eta_{i} f(e_{i}^{k})$
where $f(e_{i}^{k}) \in \mathbb{R}$ is the value of $f$ at cell $e_{i}^{k}$. The space of $k$-cochains is defined as $C^{k}(X)$, which forms a real vector space with the dual basis $\{(e_{i}^{k})^{*}\}_{i=1}^{N_{k}}$ such that $(e_{i}^{k})^{*} (e_{j}^{k}) = \delta_{ij}$.
\end{definition}
\begin{definition}
\label{def:k-coboundaryoperator}
Given definition \ref{def:k-cochain}, the coboundary operator naturally follows as $d_{k}: C^{k}(X) \to C^{k+1}(X)$ which, for $0 \leq k \leq n$, is defined as $d_{k} = f(\partial_{k+1}(c))$ for all $f \in C^{k}(X)$ and $c \in C_{k+1}(X)$. Note that for $k \in \{-1,n\}$ $d_{k}f \equiv 0$.
\end{definition}

Using these definitions, \citet{alain2023gaussian} formally introduce a generalization of the Laplacian for graphs. They further prove that for $k=0$ and identity weights, the Hodge Laplacian is the graph Laplacian. In essence, the authors prove $W_{0} = I \implies \Delta_{0} = B_{1}W_{1}B_{1}^{\top}$ and $W_{1} = I \implies \Delta_{0} = B_{1}B_{1}^{\top}$, which is the standard graph Laplacian.

\begin{definition}
\label{def:innerproduct}
Let $X$ be a finite complex. Then, we define a set of weights for every $k$. Namely, let $\{w_{i}^{k}\}_{i=1}^{N_{k}}$ be a set of real valued weights. Then, $\forall f,g \in C^{k}(X)$, one can write the weighted $L^{2}$ inner product as:$\langle f , g \rangle_{L^{2}(w^{k})} := \sum_{i=1}^{N_{k}} w_{i}^{k} f(e_{i}^{k})g(e_{i}^{k})$. This inner product induces an adjoint of the coboundary operator $d_{k}^{*}: C^{k+1}(X) \to C^{k}(X)$. Namely, $\langle d_{k}^{*} f , g \rangle = \langle f, d_{k}g \rangle$ for all $f \in C^{k+1}(X)$ and $g \in C^{k}(X)$.
\end{definition}

\begin{definition}
\label{def:hodgelaplacian}
Using the previous definitions, the Hodge Laplacian $\Delta_{k} : C^{k}(X) \to C^{k}(X)$ on the space of $k$-cochains is then $\Delta_{k} := d_{k-1} \circ d^{*}_{k-1} + d^{*}_{k} \circ d_{k}$. The matrix representation is then $\Delta_{k} := B_{k}^{\top}W_{k-1}^{-1}B_{k}W_{k} + W_{k}^{-1}B_{k+1}W_{k+1}B_{k+1}^{\top}$. Here, $W_{k} = \text{diag}(w_{1}^{k},\ldots,w_{N_{k}}^{k})$ is the diagonal matrix of cell weights and $B_{k}$ is the order $k$ incidence matrix, whose $j$-th column corresponds to a vector representation of the cell boundary $\partial e_{j}^{k}$ viewed as a $k-1$ chain.
\end{definition}

\subsection{CW-Complex Networks}
Recent advances have proposed neural networks for CW-complexes. We discuss these methods and describe key distinctions between our method. \citet{hajij2021CWC} propose an inter-cellular message-passing scheme on cell complexes. Under the proposed scheme, the propagation algorithm then performs a sequence of message passing executed between cells in $X$ defined as
\begin{equation}
    h_{c^{n-1}}^{(k)} := \alpha_{n-1}^{(k)}\left(h_{c^{n-1}}^{(k-1)}, E_{a^{n-1} \in \mathcal{N}_{adj}(c^{n-1})}\left(\phi_{n-1}^{(k)}(h_{c^{n-1}}^{(k-1)}, h_{a^{n-1}}^{(k-1)}, F_{e^{n} \in \mathcal{CO}[a^{n-1}, c^{n-1}]}(h_{e^{n}}^{(0)}))\right)\right)
\end{equation}
where $h_{e^{m}}^{(k)}$, $h_{a^{m}}^{(k)}$, $h_{c^{m}}^{(k)} \in \mathbb{R}^{\ell_{m}^{k}}$, $E$, and $F$ are permutation invariant differentiable functions and $\alpha_{j}^{(k)}$, $\phi_{j}^{(k)}$ are trainable differentiable functions. \citet{hajij2021CWC} then utilize this message passing framework to define a convolutional message passing scheme $H^{(k)} := \text{ReLU}(\hat{A}_{adj}H^{(k-1)}W^{k-1})$. We improve upon this framework in numerous ways. First we leverage the Hodge-Laplacian thereby making our network Hodge-aware. This Hodge-awareness allows for effective learning on CW-complexes when compared to message-passing networks \citep{bodnar21WLGGO}. Additionally, we rely on boundary operators $B_{k+1}^{\top}$ and $B_{k+1}$ as opposed to using message passing or permutation invariant differentiable functions. This enables a more time and space efficient way in which to process the CW-complex. We show in Lemma \ref{lemma:dimofCNNLayer} that $\dim(H^{(k)}) = N_{k} \times N_{k}$. Moreover, our form of convolution is defined for any non-linearity not just $\text{ReLU}$. Finally, as stated by \citet{hajij2021CWC}, one may wish to train a CCXN for every $k$-cells adjacency matrix individually. Consequently, one has to train $n-1$ many  networks Contrastingly, our architecture requires training only one network for every $k$-cells adjacency matrix.~\\
Similarly \citet{bodnarNeurIPS} extend a message-passing algorithm to cell complexes. \citet{bodnarNeurIPS} essentially define lifting transformations, $f: G \to X$, augmenting a graph with higher-dimensional constructs. This results in a multi-dimensional and hierarchical message passing procedure over the input graph \citep{bodnarNeurIPS}. The authors specify this procedure over the space of chemical graphs in section $4$, defining message passing from atoms to bonds, and bonds to rings. In this manuscript we propose a network that receives as input a CW-complex of dimension $n \geq 1$, which need not be a graph. Additionally we do not rely on lifting transformations or an explicit message passing scheme. In Appendix D, \citet{bodnarNeurIPS} discuss how one can define a convolutional operator on cochains. However, the authors leave development of an actual convolutional network as future work and do not mention attention \citep{bodnarNeurIPS}. We proceed in this direction by developing a Hodge-Laplacian informed convolutional network that receives a CW-complex as input.~\\
\citet{giusti23} introduce a neural architecture operating on data defined over the vertices of a graph. The approach described leverages a lifting algorithm that learns edge features from node features, then applies a cellular attention mechanism, and finally applies pooling. In particular, \citet{giusti23} define a cellular lifting map as a skeleton-preserving function $s : G \to C_{G}$ incorporating graph $G$ into regular cell complex $C_{G}$. Using the cellular lifting map the authors define attentional lift, giving way to their attention mechanism. The procedure computes $F^{0}$ many attention heads such that when given input graph $G = (\mathcal{V}, \mathcal{E})$, for vertices $i,j \in \mathcal{V}$ connected by edge $e \in \mathcal{E}$, edge features $x_{e} \in \mathbb{R}^{F^{0}}$ are computed by concatenating attention scores. Mathematically written as 
\begin{equation}
    x_{e} = g(x_i, x_j) = \mathbin\Vert_{k=1}^{F^0} a_{n}^{k}(x_i, x_j),~\forall~e \in \mathcal{E}
\end{equation}
\citet{giusti23} utilize equation (2) above to define their layer propagation scheme by combining the attentional lift with message passing and aggregating using any permutation invariant operator, see equation 6 in \citep{giusti23}. We improve on this approach in numerous ways. First we leverage the Hodge-Laplacian thereby making our network Hodge-aware. This Hodge-awareness allows for effective learning on CW-complexes when compared to message-passing networks \citep{bodnar21WLGGO}. In contrast to the work of \citet{giusti23}, our network does not rely on incorporating an input graph $G$ into a regular cell complex $C_{G}$ or attentional lift. Consequently, our method saves computations. By propagating over cells via the boundary operator, our method can account for the topology of the individual cell and its open neighborhoods. Finally, our method is leveraged to develop multi-head attention and a transformer-like architecture.

\section{Hodge-Laplacian informed CW-Complex Networks}
We extend the work of \citet{kipfwelling2017semisupervised} and \citet{GraphAttentionVelickovicBengio} by developing a notion of convolution and attention for CW-complexes that is informed by the Hodge-Laplacian. 

\subsection{Convolutional CW-complex Layer}

\begin{definition}
\label{def:CWCNN}
Let $X$ be a finite $n$-dimensional CW-complex. Then for $k \in [0, n]$ let $A_{k} \in \mathbb{R}^{N_{k} \times N_{k}}$ be a matrix of cell feature vectors and let $\Delta_{k}$ be the Hodge Laplacian. Then we can define a \textbf{Convolutional CW-complex Network (CW-CNN)} $f(X)$ as being composed by stacking hidden layers $H^{(k)}$ according to the following layer-wise propagation rule:
\begin{equation}
    H^{(k+1)} = \sigma\left( B_{k+1}^{\top}\left(\Delta_{k}A_{k} H^{(k)}\right) B_{k+1}\right)
\end{equation}
Initially, we set $H^{(0)} = X^{0} \in \mathbb{R}^{N_{0} \times N_{0}}$, which is the matrix representation of the zero-skeleton of CW-complex $X$. Recall the definitions of the boundary operator (definition \ref{def:k-boundaryoperator}), coboundary operator (definition \ref{def:k-coboundaryoperator}), and Hodge-Laplacian (definition \ref{def:hodgelaplacian}).
By definition \ref{def:hodgelaplacian}, $\Delta_{k} = B_{k}^{\top}W_{k-1}^{-1}B_{k}W_{k} + W_{k}^{-1}B_{k+1}W_{k+1}B_{k+1}^{\top}$. Additionally, by definitions \ref{def:k-boundaryoperator} and \ref{def:k-coboundaryoperator}, $B_{k} \in \mathbb{Z}^{N_{k-1} \times N_{k}}$, $B_{k+1}^{\top} \in \mathbb{Z}^{N_{k+1} \times N_{k}}$, and the weight matrix $W_{k} = \text{diag}(w_{1}^{k},\ldots,w_{N_{k}}^{k}) \in \mathbb{R}^{N_{k} \times N_{k}}$.\\
Checking the dimensions we can see that $\Delta_{k} \in \mathbb{R}^{N_{k} \times N_{k}}$:
\begin{equation}
\dim(B_{k}^{\top}W_{k-1}^{-1}B_{k}W_{k}) = (N_{k} \times N_{k-1}) (N_{k-1} \times N_{k-1}) (N_{k-1} \times N_{k}) (N_{k} \times N_{k}) = N_{k} \times N_{k}
\end{equation}
\begin{equation}
\dim(W_{k}^{-1}B_{k+1}W_{k+1}B_{k+1}^{\top}) = (N_{k} \times N_{k}) (N_{k} \times N_{k+1}) (N_{k+1} \times N_{k+1}) (N_{k+1} \times N_{k}) = N_{k} \times N_{k}
\end{equation}
Therefore $\Delta_{k} \in \mathbb{R}^{N_{k} \times N_{k}} \implies \dim(\Delta_{k} A_{k}) = N_{k} \times N_{k}$. Additionally, we know from above $\dim(B_{k+1}) = N_{k} \times N_{k+1}$ and $\dim(B_{k+1}^{\top}) = N_{k+1} \times N_{k}$. Therefore, by induction, we can show $\dim(H^{(k)}) = N_{k} \times N_{k}$ (Lemma \ref{lemma:dimofCNNLayer}).
Formally, we call $B_{k}$ the order $k$ incidence matrix, and let $\sigma$ be any nonlinearity. Thus, using the layer-wise propagation rule from equation (1), we can define a neural network $f(X)$ by stacking the hidden layers $H^{(k)}$. We call such a network a Convolutional CW-complex Network, or CW-CNN for short.
\end{definition}

\begin{lemma}
\label{lemma:dimofCNNLayer}
The dimension of hidden layer $k$ in a CW-CNN is $\dim(H^{(k)}) = N_{k} \times N_{k}$.
\begin{proof}
We want to show that $\dim(H^{(k)}) = N_{k} \times N_{k}$. For the base case ($k=0$) we define $H^{(0)} = X^{0} \in \mathbb{R}^{N_{0} \times N_{0}}$. Let the inductive hypothesis $P(j)$ be that $\forall j \in \{0,1,\ldots,k-1\}$ $\dim(H^{(j)}) = N_{j} \times N_{j}$. Then, we can show $P(j) \implies P(j+1)$ using equation (1). We know $\dim(H^{(j+1)}) = \dim(\sigma(B_{j+1}^{\top}(\Delta_{j}A_{j}H^{(j)})B_{j+1})) = (N_{j+1} \times N_{j}) (N_{j} \times N_{j}) (N_{j} \times N_{j}) (N_{j} \times N_{j}) (N_{j} \times N_{j+1}) = N_{j+1} \times N_{j+1}$. Thus we see $\dim(H^{(k)}) = N_{k} \times N_{k}$.
\end{proof}
\end{lemma}

The weight matrices $W_{k}$ can be randomly initialized by choosing $w_{i}^{k}$ randomly or by adopting a similar strategy to He Initialization \citep{HeInitializationCoRR}. In order to train this network, one would update the weight matrices $W_{k}$ with gradient descent and define a loss function $\mathcal{L}$. One can replace $W_{k}^{-1}$ by $W_{k}^{\dagger}$, the Moore-Penrose pseudoinverse of $W_{k}$, in the expression for $\Delta_{k}$. Then, let $\mathcal{D} = \{(\mathbf{x}_i, \mathbf{y}_i)\}_{i=1}^{M}$ be a dataset  where $\mathbf{x}_i$ is a CW-complex and $\mathbf{y}_{i} \in \mathbb{R}^{d}$ where $d \in \mathbb{N}$. Our learning paradigm then becomes $\mathbf{y}_{i} = f(\mathbf{x}_{i}) + \varepsilon$, where $f$ is a CW-CNN. If we let $\mathbf{X} = \left[\mathbf{x}_1, \ldots, \mathbf{x}_{M} \right]$ and $\mathbf{y} = \left[\textbf{y}_1, \ldots, \textbf{y}_{M} \right]$ we want to choose the weight matrices $W_{k}$ for each CW-complex $\textbf{x}_i$ such that we solve $\min_{W} \|f(\mathbf{X}) - \mathbf{y}\|_{2}^{2}$.

\subsection{CW-complex Attention}
\label{sec:CWCAttn}
In order to define, a CW-complex Attention Network (CW-AN) we must first develop a notion of connectedness or adjacency for CW-complexes. The analogue of adjacency for CW-complexes is termed incidence. We collect these incidence values, which are integers, in a matrix. This matrix is defined below as $B_{k}$ in equation (4).

\subsubsection{Incidence Matrices}
Let the relation $\prec$ denote incidence. If two cells $e_{j}^{k-1}$ and $e_{i}^{k}$ are incident, we write $e_{j}^{k-1} \prec e_{i}^{k}$. Similarly, if two cells $e_{j}^{k-1}$ and $e_{i}^{k}$ are not incident, we write $e_{j}^{k-1} \nprec e_{i}^{k}$. Additionally, let the relation $\sim$ denote orientation. We write $e_{j}^{k-1} \sim e_{i}^{k}$ if the cells have the same orientation. If two cells have the opposite orientation we write $e_{j}^{k-1} \nsim e_{i}^{k}$. This enables us to define the classical incidence matrix \citep{IncidenceMatrix}. Traditionally we define for indices $i$ and $j$ the value of $(B_{k})_{j,i}$ as:
\begin{equation}
    (B_{k})_{j,i} = \begin{cases} 0 ~~~~\text{if}~e_{j}^{k-1} \nprec e_{i}^{k} \\ 1 ~~~~\text{if}~e_{j}^{k-1} \prec e_{i}^{k}~\text{and}~ e_{j}^{k-1} \sim e_{i}^{k} \\ -1 ~~\text{if}~e_{j}^{k-1} \prec e_{i}^{k}~\text{and}~ e_{j}^{k-1} \nsim e_{i}^{k} \end{cases}
\end{equation}
The matrix $B_{k}$ establishes which $k$-cells are incident to which $k-1$-cells. We know from above that $B_{k} \in \mathbb{Z}^{N_{k-1} \times N_{k}}$. Let $\Col_{j}(B_{k})$ denote the $j$-th column of $B_{k}$. We know that $\Col_{j}(B_{k}) = \mathbb{Z}^{N_{k-1} \times 1}$ which corresponds to a vector representation of the cell boundary $\partial e_{j}^{k}$ viewed as a $k-1$ chain \citep{alain2023gaussian}.

Using the definition of an incidence matrix and the relation $\prec$ we can now define a CW-AN.

\begin{definition}
\label{def:CWAT}
Let $X$ be a finite $n$-dimensional CW-complex. Then for $k \in [0, n]$ we can define a \textbf{CW-complex Attention Network (CW-AN)}. A CW-AN computes a learned weighted average of the representations of each skeleton. We start by initializing for all $i$ the starting hidden state\\ $h_{e_{i}^{0}}^{(0)} = \begin{bmatrix} e_{1}^{0} , \ldots, e_{N_0}^{0} \end{bmatrix}^{\top} \left( \Col_{i}(B_{0}^{\top})\right)^{\top} \in \mathbb{R}^{N_{0} \times N_{0}}$. We then define a scoring function $\mathcal{S}$
\begin{equation}
    \mathcal{S}(h_{e_{i}^{k}}^{(k)},h_{e_{j}^{k-1}}^{(k-1)}) = \text{LeakyReLU}\left(\left[ W_{k}h_{e_{i}^{k}}^{(k)}  \mathbin\Vert B_{k}^{\top} W_{k-1}h_{e_{j}^{k-1}}^{(k-1)} B_{k}  \right] a^{\top} \right) 
\end{equation}
where $W_{k} \in \mathbb{R}^{N_{k} \times N_{k}}$, $W_{k-1} \in \mathbb{R}^{N_{k-1} \times N_{k-1}}$, $B_{k} \in \mathbb{R}^{N_{k-1} \times N_{k}}$, $a^{\top} \in \mathbb{R}^{2 N_{k} \times N_{k}}$, and $\mathbin\Vert$ denotes vector concatenation. Checking dimensions, we see that:
\begin{equation}
    \dim(W_{k}h_{e_{i}^{k}}^{(k)}) = (N_{k} \times N_{k}) (N_{k} \times N_{k}) = N_{k} \times N_{k}
\end{equation}
Additionally, for $e_{j}^{k-1}$ we see:
\begin{equation}
    \dim(B_{k}^{\top}W_{k-1}h_{e_{j}^{k-1}}^{(k-1)}B_{k}) = (N_{k} \times N_{k-1}) (N_{k-1} \times N_{k-1}) (N_{k-1} \times N_{k-1}) (N_{k-1} \times N_{k}) = N_{k} \times N_{k}
\end{equation}
By equations (6) and (7) we know:
\begin{equation}
\dim\left(\left[W_{k}h_{e_{i}^{k}}^{(k)} \mathbin\Vert B_{k}^{\top} W_{k-1}h_{e_{j}^{k-1}}^{(k-1)} B_{k}  \right]\right) = N_{k} \times 2N_{k}
\end{equation}
Equations (5) and (8) imply:
\begin{equation}
\dim\left( \mathcal{S}(h_{e_{i}^{k}}^{(k)},h_{e_{j}^{k-1}}^{(k-1)}) \right) = 
\dim\left( \text{LeakyReLU}\left(\left[ W_{k}h_{e_{i}^{k}}^{(k)} \mathbin\Vert B_{k}^{\top} W_{k-1}h_{e_{j}^{k-1}}^{(k-1)} B_{k}  \right] a^{\top} \right)  \right) = N_{k} \times N_{k}
\end{equation}
Therefore, we can compute attention scores $\alpha_{e_{i}^{k}, e_{j}^{k}}$, which are normalized over all incident cells. Let the set $\mathcal{N}_{e_{i}^{k}} := \left\{ e_{j'}^{k-1} ~|~ e_{j'}^{k-1} \prec e_{i}^{k} \land (e_{j'}^{k-1} \sim e_{i}^{k} \lor e_{j'}^{k-1} \nsim e_{i}^{k}) \right\}$ contain all cells incident to $e_{i}^{k}$. We define a variation of the standard Softmax function called $\CSoftmax$, which converts cells to a probability distribution.
\begin{equation}
\alpha_{e_{i}^{k}, e_{j}^{k-1}} = \CSoftmax\left(\mathcal{S}(h_{e_{i}^{k}}^{(k)},h_{e_{j}^{k-1}}^{(k-1)}) \right) = \frac{\exp\left( \mathcal{S}(h_{e_{i}^{k}}^{(k)},h_{e_{j}^{k-1}}^{(k-1)}) \right)}{ \mathlarger{\sum}_{e_{j'}^{k-1} \in \mathcal{N}_{e_{i}^{k}}} \exp\left( \mathcal{S}(h_{e_{i}^{k}}^{(k)},  h_{e_{j'}^{k-1}}^{(k-1)})   \right)  } \in \mathbb{R}^{N_{k} \times N_{k}} 
\end{equation}
This enables us to define our update rule for computing later hidden states:
\begin{equation}
    h_{e_{i}^{k}}^{(k)} = \sigma\left( \mathlarger{\sum}_{e_{j'}^{k-1} \in \mathcal{N}_{e_{i}^{k}}} \alpha_{e_{i}^{k}, e_{j'}^{k-1}}  B_{k}^{\top} W_{k-1} h_{e_{j'}^{k-1}}^{(k-1)} B_{k} \right)
\end{equation}
We know from equation (7) that
\begin{equation}
    \dim(h_{e_{i}^{k}}^{(k)}) =  \dim\left( \alpha_{e_{i}^{k}, e_{j'}^{k-1}}  B_{k}^{\top} W_{k-1} h_{e_{j'}^{k-1}}^{(k-1)} B_{k} \right) = (N_{k} \times N_{k}) (N_{k} \times N_{k}) = N_{k} \times N_{k}
\end{equation}

Thus, we have defined a notion of  self-attention. A CW-AN is then composed by stacking numerous $h_{e_{i}^{k}}^{(k)}$ for all $k \in [0,n]$ and all cells.
\end{definition}

We note that all $W_{i}$ and $a^{\top}$ are learned. To stabilize the learning process of this self-attention mechanism, we can extend definition \ref{def:CWAT} to develop multi-head attention. Just as in \citet{GraphAttentionVelickovicBengio}, we can concatenate $K$ independent self-attention mechanisms to execute equation (11). This results in the following representation
\begin{equation}
    h_{i}^{'} = \displaystyle{\concat_{\ell=1}^{K}} \sigma\left( \mathlarger{\sum}_{e_{j'}^{k-1} \in \mathcal{N}_{e_{i}^{k}}} \alpha_{e_{i}^{k}, e_{j'}^{k-1}}^{(\ell)}  B_{k}^{\top} W_{k-1}^{(\ell)} h_{e_{j'}^{k-1}}^{(k-1)} B_{k} \right)
\end{equation}
where $\alpha_{e_{i}^{k}, e_{j'}^{k-1}}^{(\ell)}$ are normalized attention coefficients computed by the $\ell$-th attention mechanism and $W_{k-1}^{(\ell)}$ is the corresponding weight matrix. In practice, the operation of the self-attention layer described in definition \ref{def:CWAT} can be parallelized. One can develop a transformer based architecture by combining the multi-head attention layer described in equation (13) with modified Add$\&$Layer Norm as well as Feed Forward networks, equivalent to those developed by \citet{ATTNALLUNEED}. The weight matrices $W_{k}$ can be randomly initialized or one may adopt a different strategy.

\section{Model Architecture}
We develop two distinct networks for one synthetic task.

\subsection{CW-CNN architecture}
Graph convolutional networks are thought of as models which efficiently propagate information on graphs \citep{kipfwelling2017semisupervised}. Convolutional networks traditionally stack multiple convolutional, pooling, and fully-connected layers to encode image-specific features for image-focused tasks \citep{CNNPaperNash}. The CW-CNN is motivated by similar principles. In particular, a CW-CNN efficiently propagates information on CW-complexes and encodes topological features for tasks in which geometry plays a pivotal role.

There are numerous potential use cases for such an architecture. In the areas of drug design, molecular modeling, and protein informatics the three dimensional structure of molecules plays a pivotal role and numerous approaches in these areas have attempted to meaningfully extract or utilize geometric information \citep{3DContrastive, molDesign, molGen, structDesign}.

A CW-CNN is composed of a stack of Convolutional CW-complex layers as described above in definition \ref{def:CWCNN}. One may additionally add pooling, linear, or dropout layers. In our experiment we stack Convolutional CW-complex layers, and follow up with a Linear layer and GELU activation. The architecture is pictured below.

\begin{figure}[h]
\begin{center}
\fbox{\rule[-.5cm]{0cm}{4cm}
\includegraphics[scale=0.35]{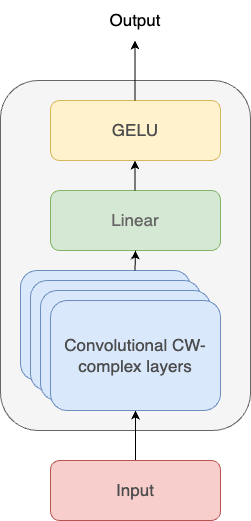}
\rule[-.5cm]{0cm}{4cm}}
\end{center}
\caption{CW-CNN architecture.}
\end{figure}

\subsection{CW-AT architecture}
Initially, Graph Attention Networks were developed as a method to perform node classification on graph-structured data \citep{GraphAttentionVelickovicBengio}. Transformers are neural sequence transduction models that maintain an encoder decoder structure, wherein an input sequence $(x_1, \ldots, x_n)$ is mapped to a sequence of continuous representations $(z_1, \ldots, z_n)$. Then $(z_1, \ldots, z_n)$  are passed to the decoder for generating an output sequence $(y_1, \ldots, y_m)$ in an auto-regressive fashion \citep{ATTNALLUNEED}. The CW-AT is motivated by similar principles. In particular, a CW-AT leverages a different kind of attention mechanism, which can perform classification or regression on CW-complex structured data. While one can theoretically setup a sequence-to-sequence learning problem utilizing CW-complexes, we do not venture into such problems. However, we develop an architecture that vaguely resembles the classic Transformer with far fewer parameters.

There are numerous potential use cases for such an architecture. One can attempt language translation, image captioning and sequence transformation tasks \citep{sutskever2014sequence}. Doing so would require viewing a word or entity as a CW-complex. This is somewhat reasonable since the $1$-skeleton of a CW-complex is a graph without loops \citep{whitehead1949combinatorial}. Graphs have appeared in numerous contexts within Natural language processing. Classically vertices can encode text units of various sizes and characteristics such as words, collocations, word senses, sentences and documents \citep{NLP}. Edges may represent relationships such as co-occurrence \citep{NLP}. One can replace the notion of vertex with cell and edge with a kind of gluing map and extend these ideas to CW-complexes. One can represent co-occurence for instance by scaling the matrix $B_{k}$.

A CW-AT is composed of two separate networks, with each receiving as input a CW-complex. In each network the input is processed by a Multi-Cellular Attention mechanism as described by equation (13). Afterwards one may apply dropout, feed forward, and layer norm. The outputs from the first network are combined with output from the second network through addition and applying SELU activation. Finally, a linear layer is used to get the desired output shape. One may apply a Softmax if the output is to be viewed as probabilities.

\begin{figure}[h]
\label{fig:CWAT}
\begin{center}
\fbox{
\rule[-.5cm]{0cm}{4cm} 
\includegraphics[scale=0.35]{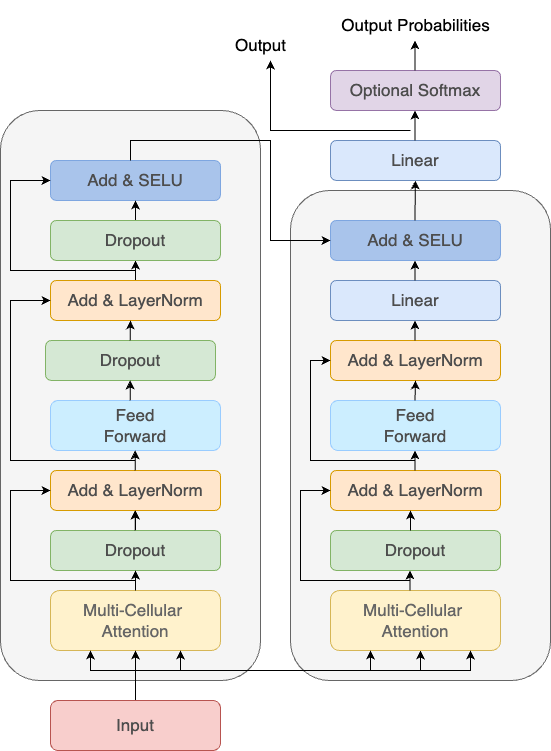}
\rule[-.5cm]{0cm}{4cm}}
\end{center}
\caption{CW-AT architecture.}
\end{figure}

\section{Experiment}

This section describes our experimental setup and training regime.

\subsection{Dataset}
We construct a synthetic dataset consisting of CW-complexes and attempt to predict the number of cells in each complex. This task is intended to be a proof of concept. Let $\mathcal{D} = \{x_{i}, y_{i}\}_{i=1}^{N}$ be our dataset. Here we choose $N = 500$ and let $x_{i}$ be a CW-complex and let $y_{i} \in \mathbb{R}$ be the number of cells found in complex $x_i$.

\subsection{Hardware and Wall clock time}
We train both of our models on one AWS m7g.2xlarge instance. For both models, each training step took approximately $0.1$ seconds. We trained both models for only $400$ steps. Therefore, total training took approximately $40$ seconds for each model.

\subsection{Optimizer}
We utilized SGD in order to train both models. For our CW-CNN we utilized learning rate $\eta = 0.001$ and momentum $\alpha = 0.9$. For our CW-AT we utilized learning rate $\eta = 0.001$, momentum $\alpha = 0.7$ and weight decay $\lambda = 0.02$.

\subsection{Regularization}
We utilized regularization during training the CW-AT. We apply dropout in the CW-AT with $P_{drop} = 0.1$ for every dropout layer shown in Figure 2.

\section{Results}
Using an 80/20 train/test split we report test set accuracy for both models. We summarize our results in Table \ref{table-1}.

\begin{table}[t]
\caption{CW-CNN and CW-AT attain low test-set RMSE.}
\label{table-1}
\begin{center}
\begin{tabular}{lll}
\multicolumn{1}{c}{\bf Model}  &\multicolumn{1}{c}{\bf RMSE} & \multicolumn{1}{c}{\bf Parameters}
\\ \hline \\
CW-AT & $0.02533091887133196$ & 310\\
CW-CNN & $1.1487752999528311 \times 10^{-5}$ & 30\\
\hline
\end{tabular}
\end{center}
\end{table}

On our synthetic task, we demonstrate low test-set RMSE. At the time of experiment, there are no other existing neural networks with which to draw comparison. We interpret our experiment as a proof of concept demonstrating that such an architecture is feasible, compute-efficient, and well-defined. In the event that one wishes to develop deeper architectures, stacking more layers or increasing model dimensions are well defined.

\section{Conclusion}

In this work, we presented the CW-CNN and CW-AT, the first types of neural network that can receive CW-complexes as input. We demonstrate high accuracy with relatively few parameters on a synthetic task. These results have implications for machine learning tasks in which geometric information or three dimensional structure plays a pivotal role. These areas include, but are not limited to, molecular design, cheminformatics and drug discovery. Additionally one may view tasks involving graphs in natural language as good candidates for a CW-complex representation and correspondingly a CW-AT. CW-complexes capture interactions between higher-order cells enabling the one to model polyadic relations effectively. Our neural networks enable learning on CW-complexes thereby facilitating the learning of polyadic relations. We are excited about the future of these models and plan to apply them to other tasks in cheminformatics and natural language processing.




\bibliography{main}
\bibliographystyle{tmlr}

\end{document}